\documentclass[letterpaper, 10 pt, journal, twoside]{IEEEtran}  

\usepackage{times}
\usepackage{graphicx} 
\usepackage[hidelinks,bookmarks=true]{hyperref}

\usepackage{cite}

\usepackage{multicol}
\usepackage{caption}

\usepackage{amsmath}
\usepackage{cuted} 
\usepackage{booktabs, array} 
\renewcommand{\arraystretch}{1.4}
\usepackage[linesnumbered,ruled,vlined]{algorithm2e}
\usepackage{pgfplots}
\usepackage{subcaption}
\pgfplotsset{compat=1.18}
\usepackage{adjustbox}
\usepackage{multirow}
\usepackage{balance}

\newif\ifshowrevisions
\showrevisionsfalse   

\ifshowrevisions
  \newcommand{\rev}[1]{{\color{blue}#1}}
\else
  \newcommand{\rev}[1]{#1}
\fi

\begin{document}

\title{
GeNIE: A Generalizable Navigation System for In-the-Wild Environments
}


\author{%
Jiaming Wang$^{1\dagger}$, Diwen Liu$^{1\dagger}$, Jizhuo Chen$^{1\dagger}$, Jiaxuan Da$^{1}$,\\
Nuowen Qian$^{1}$, Minh Man Tram$^{1}$, and Harold Soh$^{1,2}$%

\thanks{$^{1}$Jiaming Wang, Diwen Liu, Jizhuo Chen, Jiaxuan Da, Nuowen Qian, Minh Man Tram, and Harold Soh are with the Department of Computer Science, School of Computing, National University of Singapore, Singapore. (email: {\tt\footnotesize jiaming@comp.nus.edu.sg})}
\thanks{$\dagger$ These authors contributed equally to this work.}
\thanks{$^{2}$Harold Soh is also with the Smart Systems Institute, National University of Singapore (email: {\tt\footnotesize harold@nus.edu.sg})}
}




\markboth{Author’s accepted manuscript (AAM) — accepted to IEEE Robotics and Automation Letters (RAL), 2025}{}

\IEEEaftertitletext{\vspace{-5.0\baselineskip}} 
\maketitle
\pagestyle{plain}
\begin{strip}
\centering
\setlength{\tabcolsep}{0pt}
\renewcommand{\arraystretch}{0}
\begin{tabular}{ccccccc}
\adjustbox{width=0.14\linewidth, height=0.08\linewidth, clip}{\includegraphics{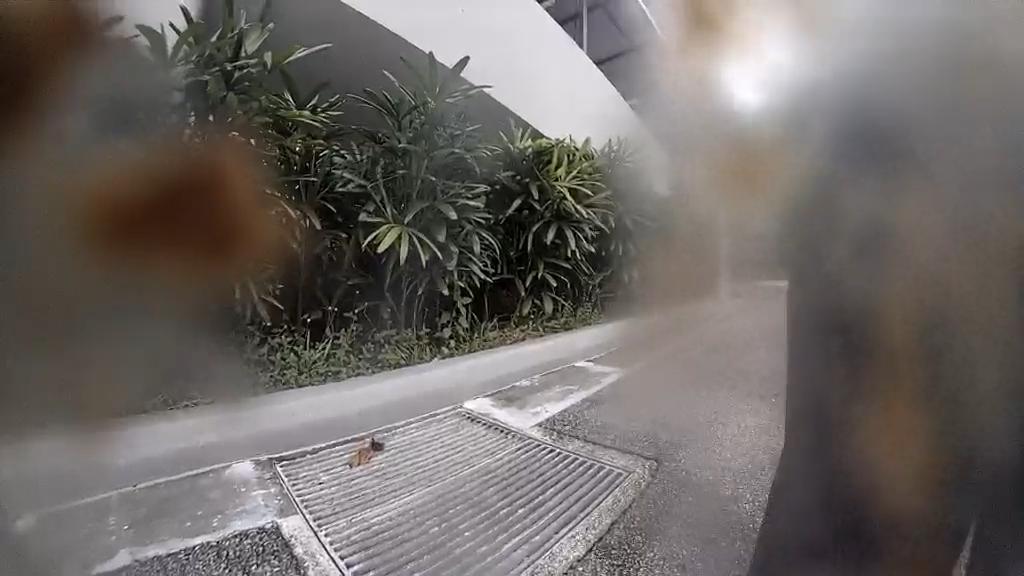}} &
\adjustbox{width=0.14\linewidth, height=0.08\linewidth, clip}{\includegraphics{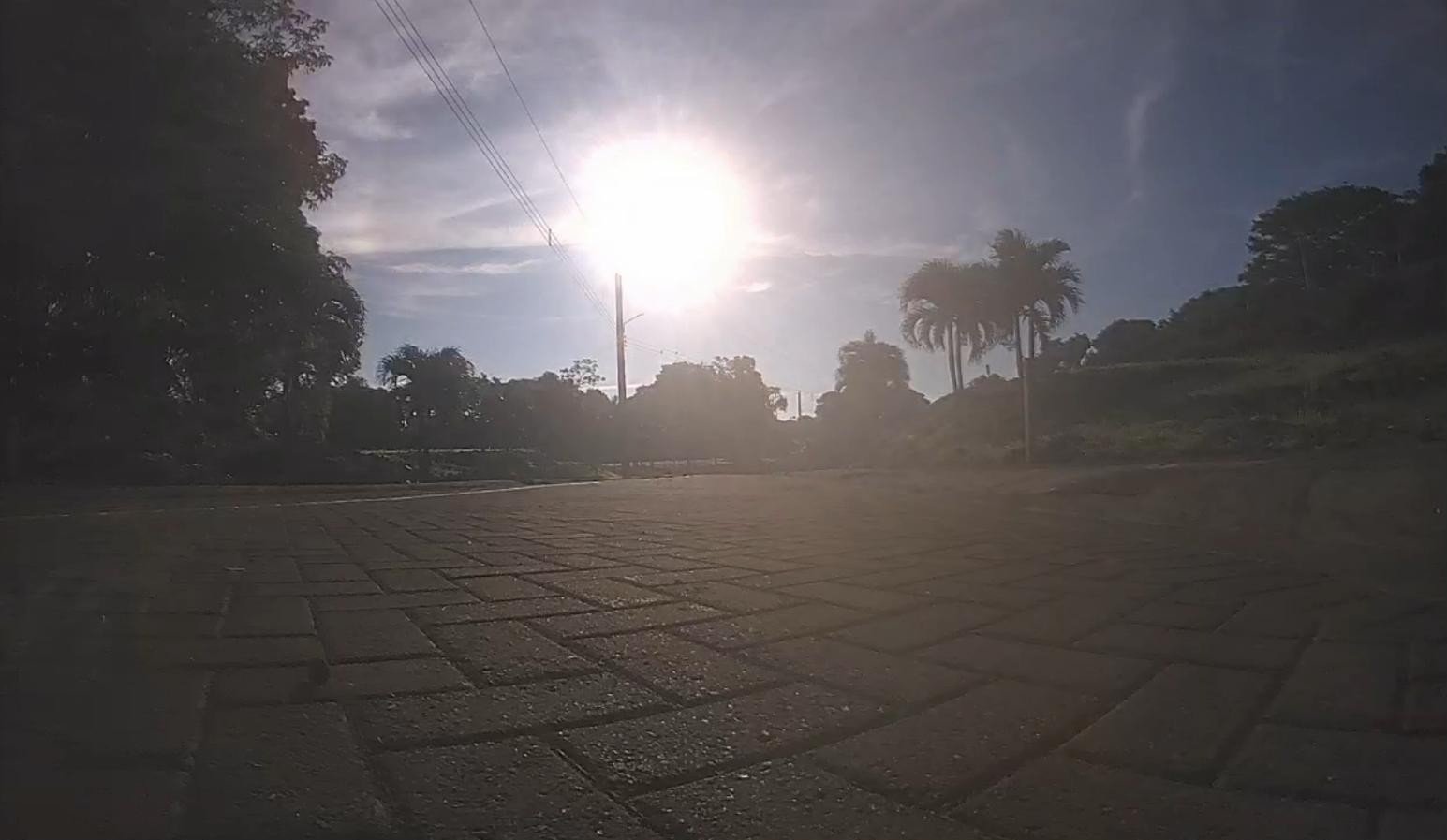}} &
\adjustbox{width=0.14\linewidth, height=0.08\linewidth, clip}{\includegraphics{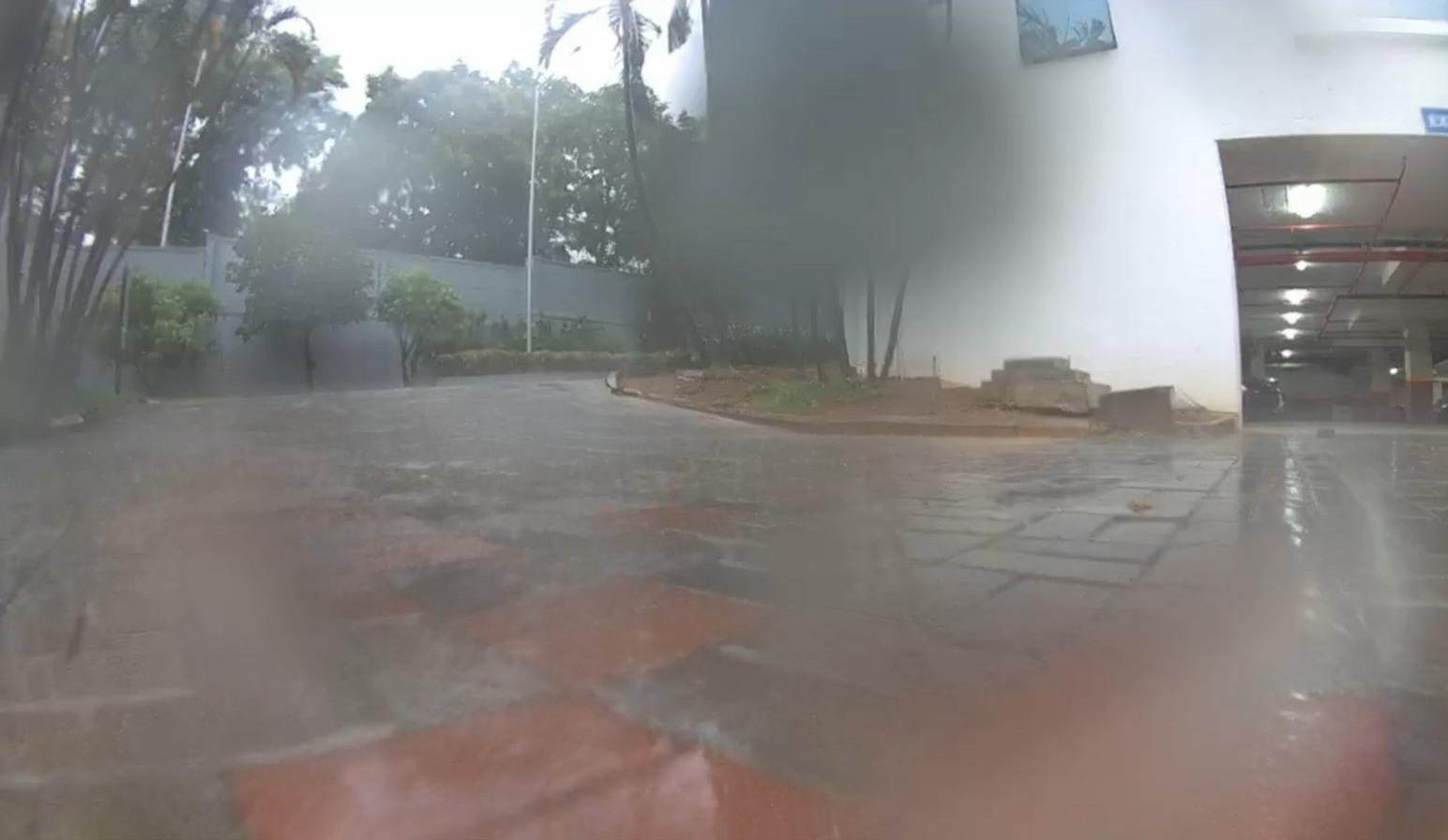}} &
\adjustbox{width=0.14\linewidth, height=0.08\linewidth, clip}{\includegraphics{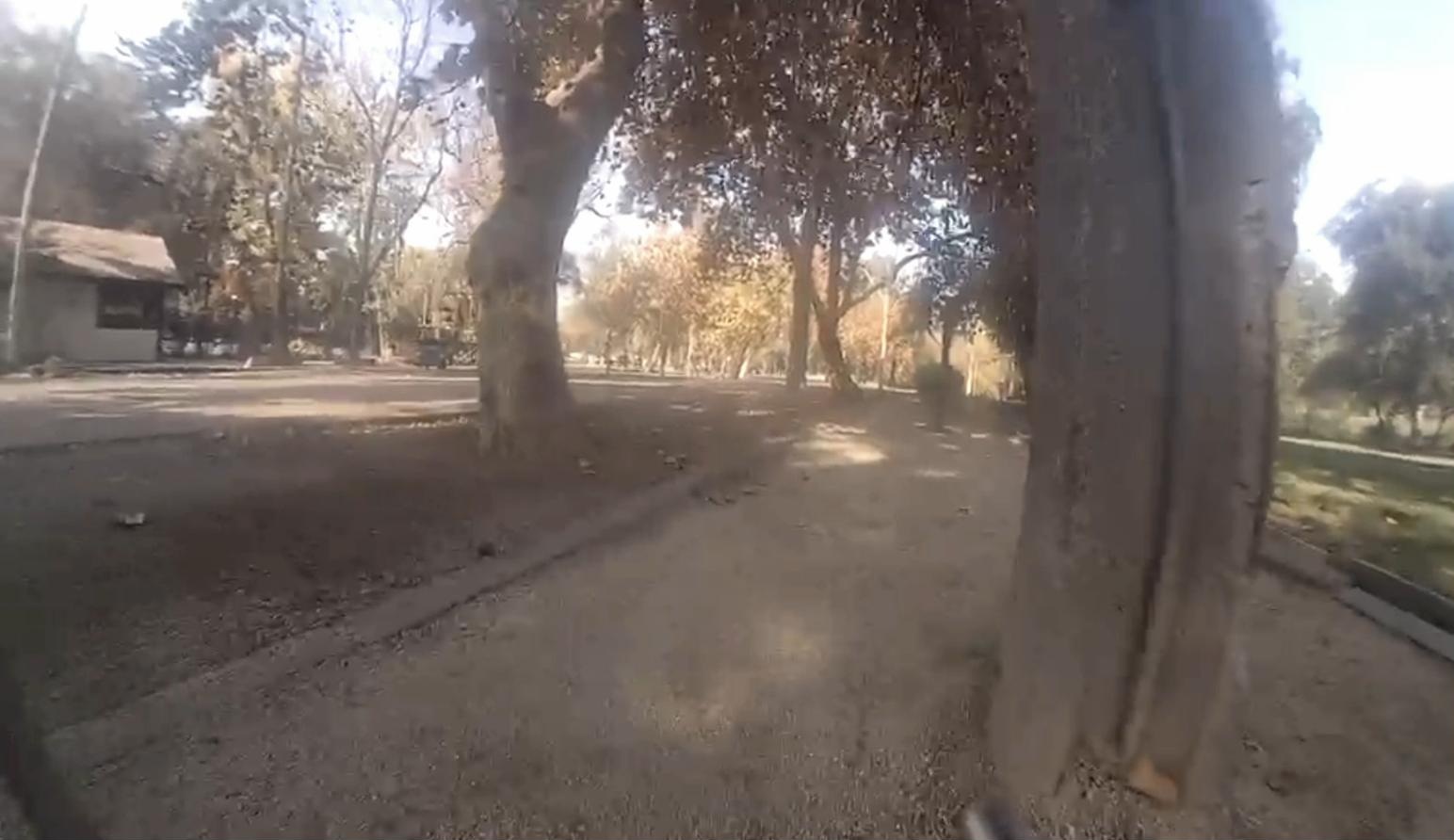}} &
\adjustbox{width=0.14\linewidth, height=0.08\linewidth, clip}{\includegraphics{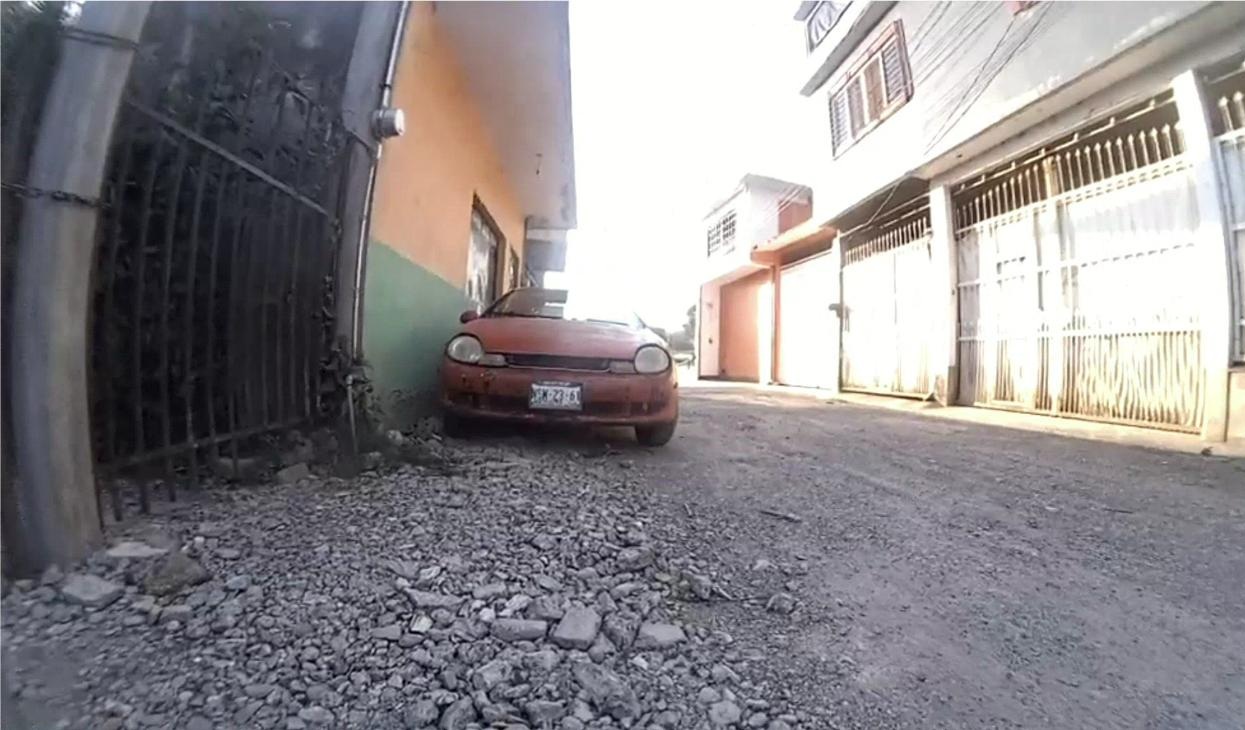}} &
\adjustbox{width=0.14\linewidth, height=0.08\linewidth, clip}{\includegraphics{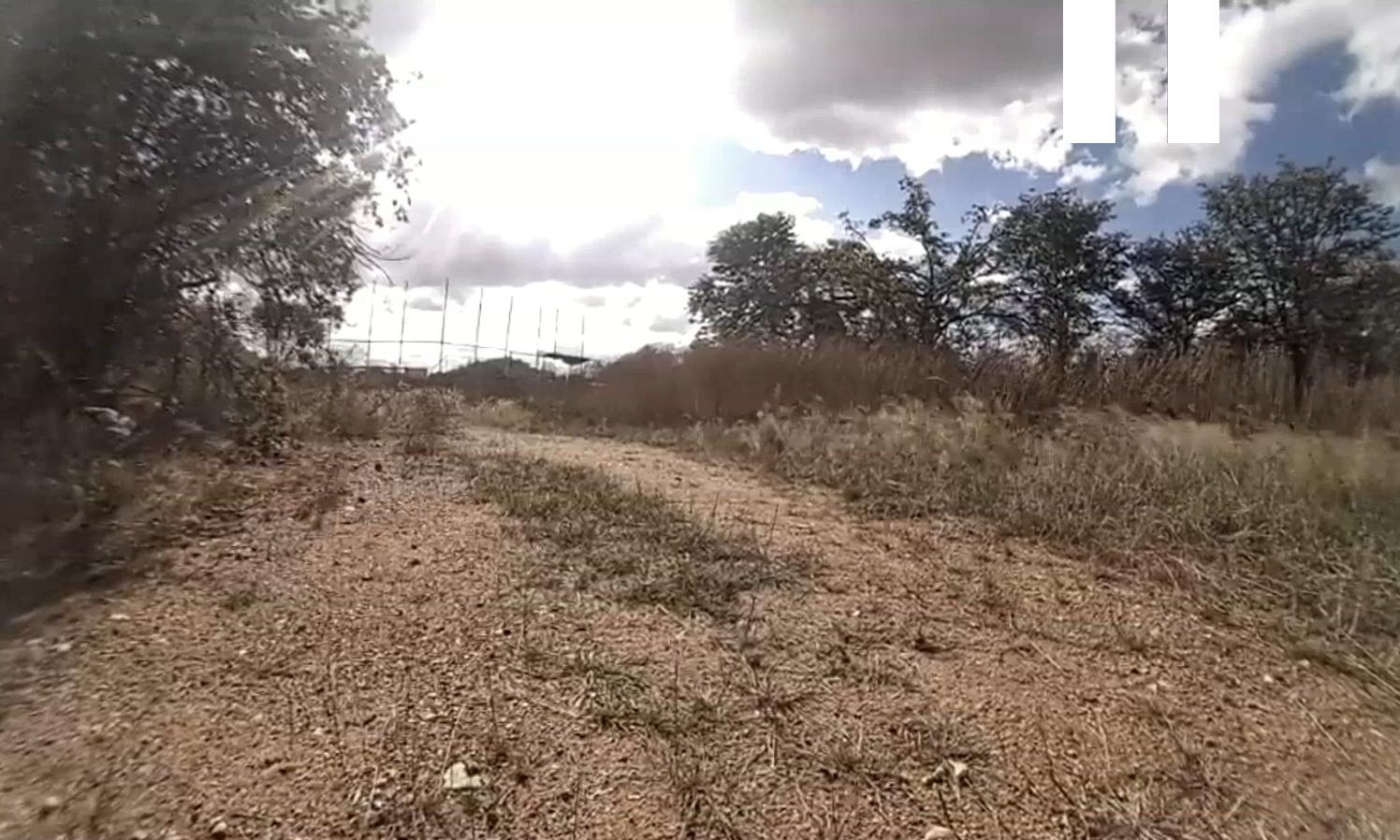}} &
\adjustbox{width=0.14\linewidth, height=0.08\linewidth, clip}{\includegraphics{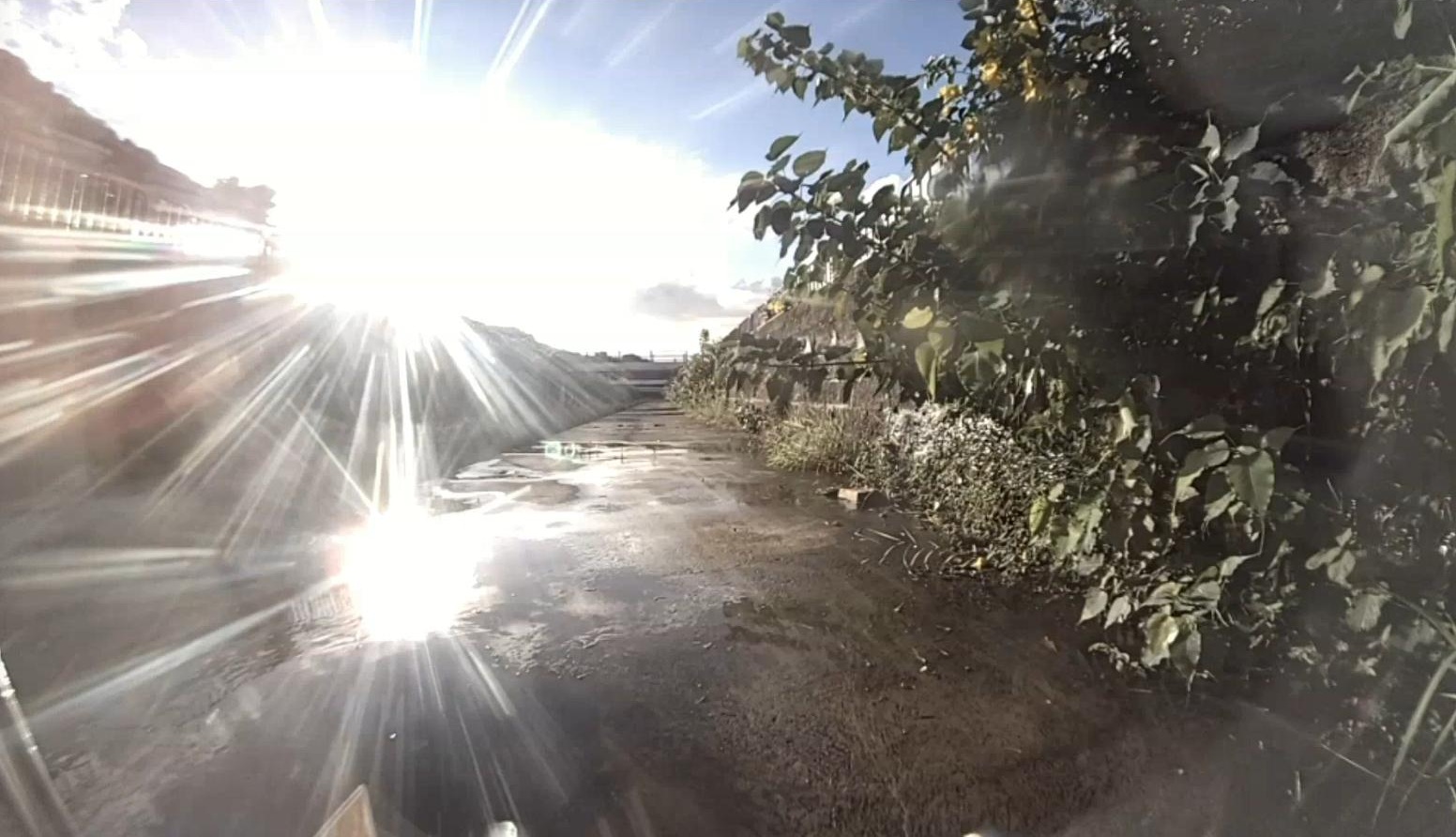}} \\
\adjustbox{width=0.14\linewidth, height=0.08\linewidth, clip}{\includegraphics{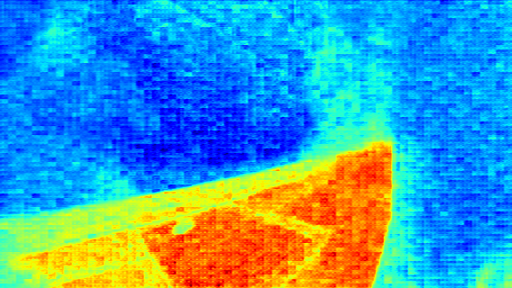}} &
\adjustbox{width=0.14\linewidth, height=0.08\linewidth, clip}{\includegraphics{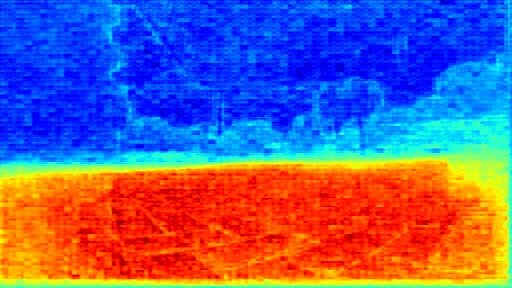}} &
\adjustbox{width=0.14\linewidth, height=0.08\linewidth, clip}{\includegraphics{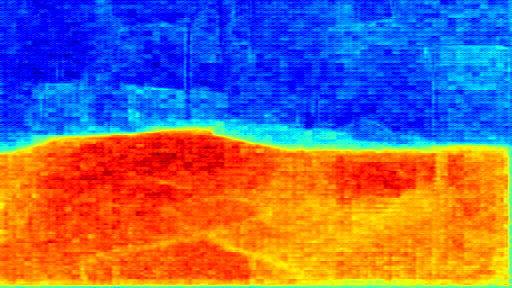}} &
\adjustbox{width=0.14\linewidth, height=0.08\linewidth, clip}{\includegraphics{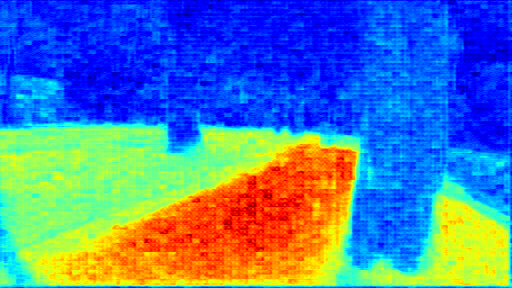}} &
\adjustbox{width=0.14\linewidth, height=0.08\linewidth, clip}{\includegraphics{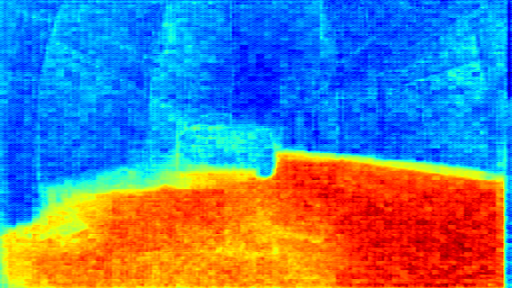}} &
\adjustbox{width=0.14\linewidth, height=0.08\linewidth, clip}{\includegraphics{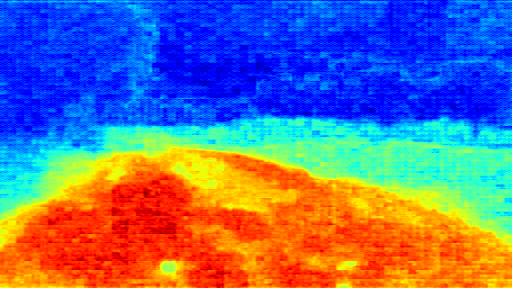}} &
\adjustbox{width=0.14\linewidth, height=0.08\linewidth, clip}{\includegraphics{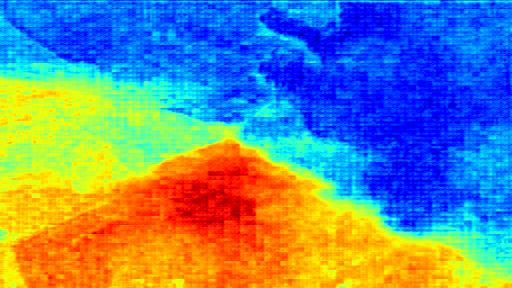}} \\
\adjustbox{width=0.14\linewidth, height=0.14\linewidth, clip}{\includegraphics{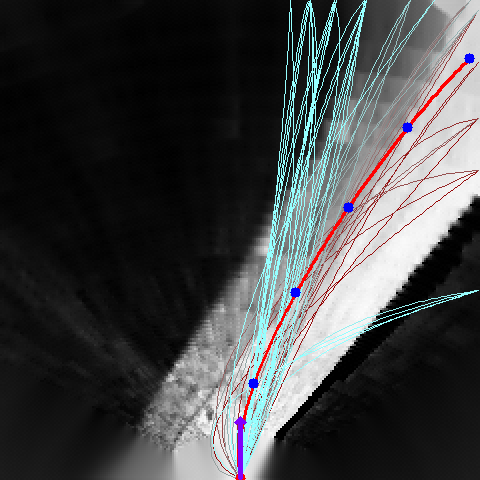}} &
\adjustbox{width=0.14\linewidth, height=0.14\linewidth, clip}{\includegraphics{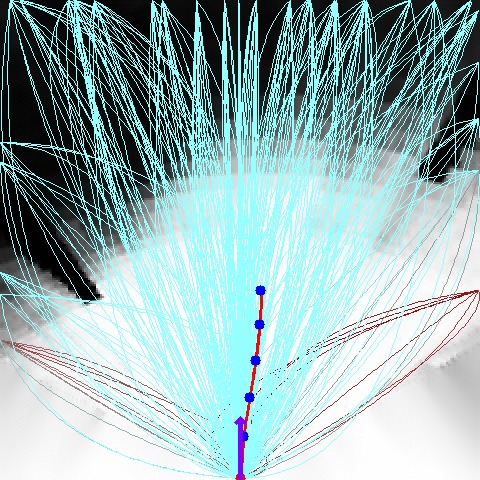}} &
\adjustbox{width=0.14\linewidth, height=0.14\linewidth, clip}{\includegraphics{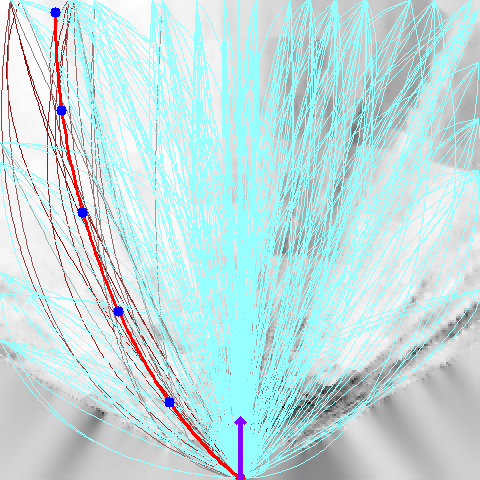}} &
\adjustbox{width=0.14\linewidth, height=0.14\linewidth, clip}{\includegraphics{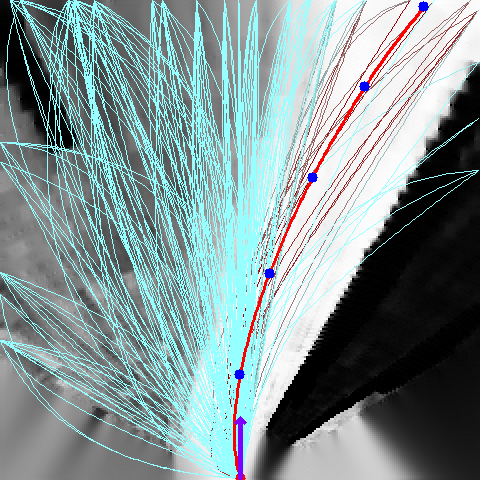}} &
\adjustbox{width=0.14\linewidth, height=0.14\linewidth, clip}{\includegraphics{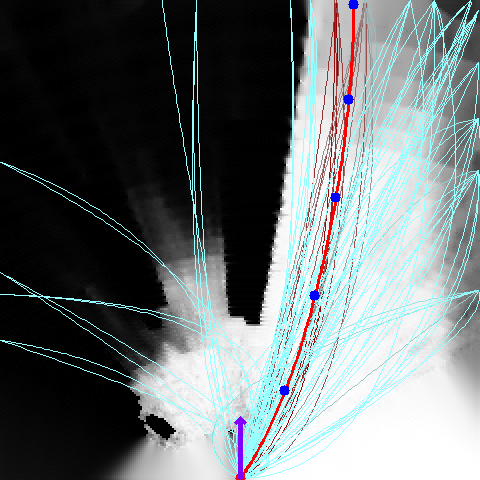}} &
\adjustbox{width=0.14\linewidth, height=0.14\linewidth, clip}{\includegraphics{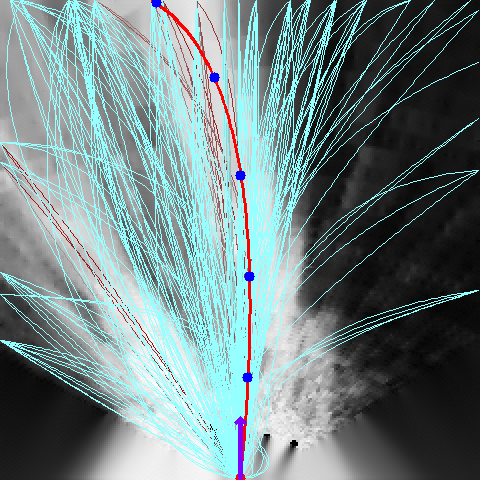}} &
\adjustbox{width=0.14\linewidth, height=0.14\linewidth, clip}{\includegraphics{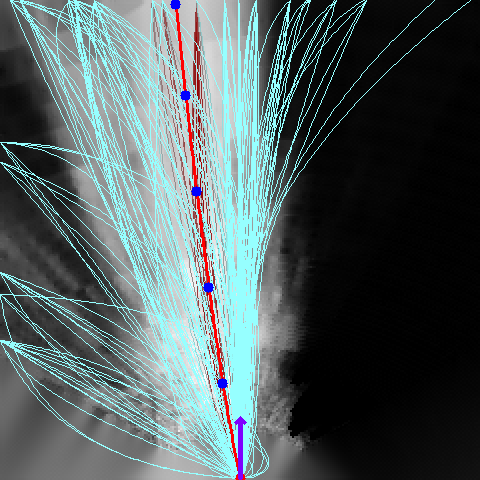}} \\
\end{tabular}
\captionof{figure}{Diverse input scenes (top), traversability predictions (middle), and final path planning results (bottom) across \rev{test environments unseen during training.} GeNIE demonstrates strong generalization under challenging conditions, including heavy rain, high-contrast lighting, muddy lenses, and diverse terrain types.}

\label{fig:diverse-scene}
\end{strip}

\begin{abstract}
Reliable navigation in unstructured, real-world environments remains a significant challenge for embodied agents, especially when operating across diverse terrains, weather conditions, and sensor configurations. In this paper, we introduce \textbf{GeNIE} (Generalizable Navigation System for In-the-Wild Environments), a robust navigation framework designed for global deployment. GeNIE integrates a generalizable traversability prediction model with a novel path fusion strategy that enhances planning stability in noisy and ambiguous settings.
We deployed GeNIE in the Earth Rover Challenge (ERC) at ICRA 2025, where it was evaluated across six countries spanning three continents. GeNIE took first place and achieved \textbf{79\% of the maximum possible score}, outperforming the second-best team by \textbf{17\%}, and completed the entire competition without a single human intervention. These results set a new benchmark for robust, generalizable outdoor robot navigation. We have released the codebase, pretrained model weights, and the datasets at \url{https://clear-nus.github.io/genie/} to support future research in real-world navigation.
\end{abstract}

\begin{figure}[t]
    \centering
    \includegraphics[width=8.9cm]{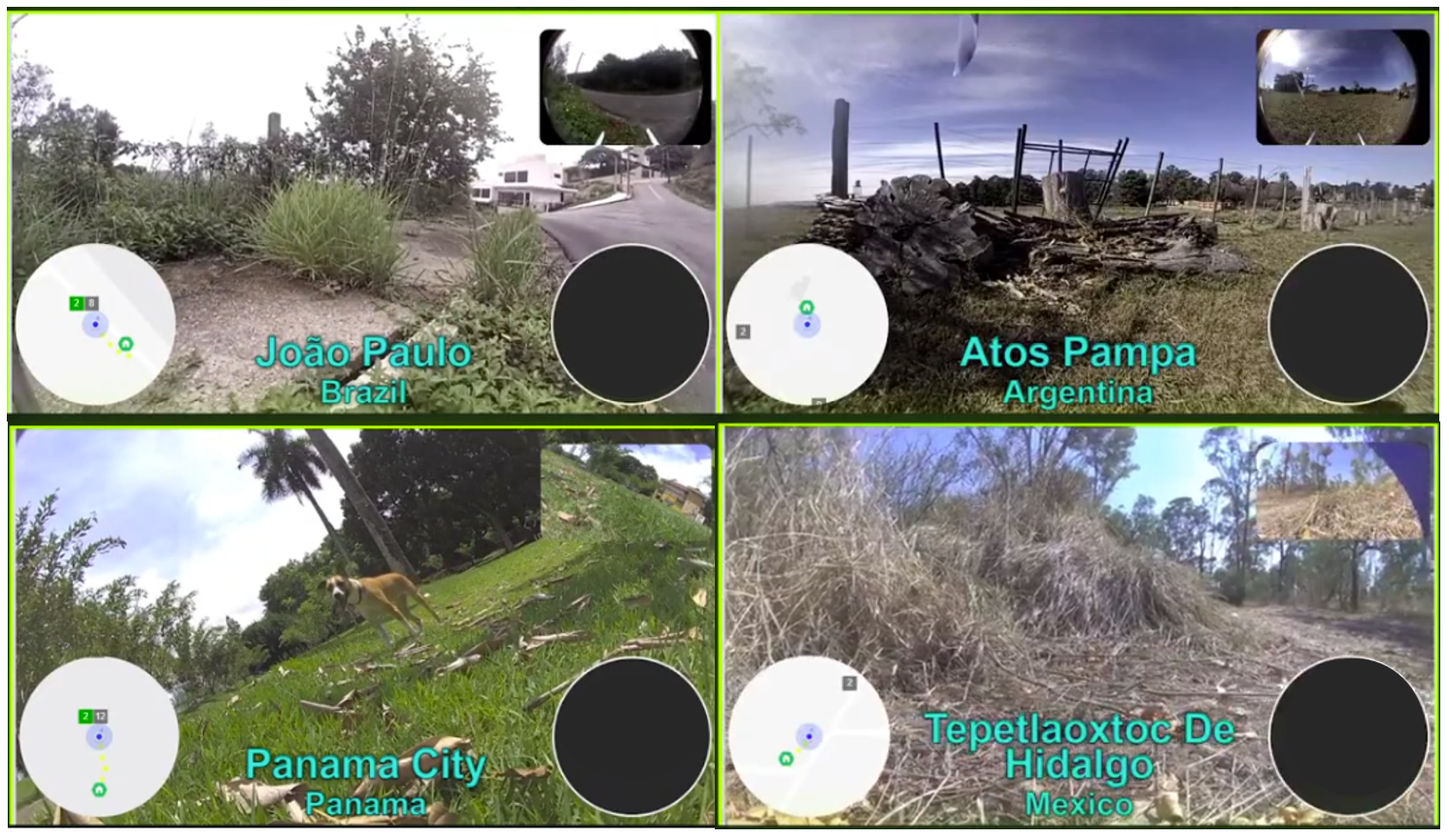}
    \caption{\rev{Examples of diverse terrains in ERC 2025, illustrating the requirement for robots to identify traversable areas in unstructured environments~\cite{erc2025_video}. Reproduced with permission from the ERC organizers.}}
    \label{fig:erc_scenes}
    \vspace{-1em}
\end{figure}

\begin{figure*}[t]
    \centering
    \includegraphics[width=0.98\linewidth]{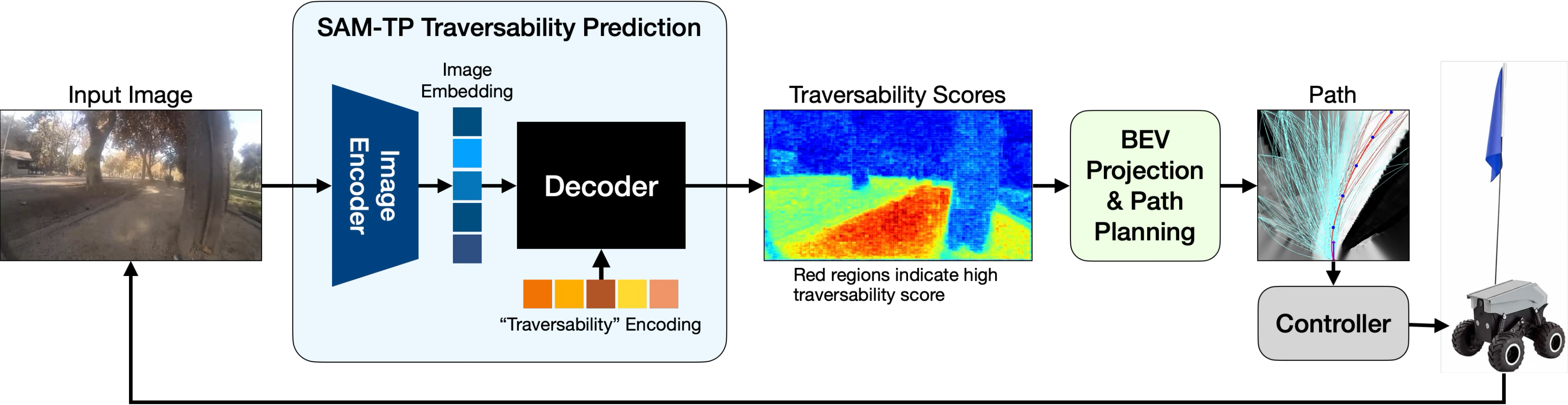}
    \caption{Overview of the GeNIE system. Given an RGB input image, the SAM-TP module predicts navigable regions in the image space. These predictions are projected into a bird's-eye view (BEV) cost map. Path fusion is then performed to identify coherent and safe traversable paths, followed by path selection based on alignment with the goal direction. Finally, the control module outputs linear and angular velocities to follow the selected path.}
    \label{fig:system_overview}
    \vspace{-1em}
\end{figure*}

\section{Introduction}
\label{sec:introduction}
\IEEEPARstart{N}{AVIGATION} is a core skill for embodied agents, allowing them to move purposefully through physical environments using inputs from onboard sensors such as cameras and GPS. It enables robots to reach goals, avoid obstacles, and operate autonomously in tasks like package delivery, environmental monitoring, or search and rescue. Over the past several years, numerous benchmarks and competitions have been introduced to evaluate navigation systems, including indoor simulation platforms such as Habitat~\cite{savva2019habitat} and Gibson~\cite{xia2018gibson}, as well as outdoor field-based challenges focused on specific domains like subterranean navigation or agricultural environments. While these efforts have significantly advanced the field, most benchmarks evaluate navigation within narrow settings, often assuming a single type of scene, weather condition, or embodiment. As a result, high scores on these benchmarks often fail to generalize when deployed in more diverse or unstructured environments~\cite{ramakrishnan2021hm3d, xia2023generalization}.

In real-world scenarios, however, navigation systems must operate reliably across a wide variety of environments, including different terrains, lighting conditions, and sensor configurations, as illustrated in Figure~\ref{fig:diverse-scene}. Domain shifts such as changing weather, unexpected obstacles, or cultural variations in street layout can easily degrade system performance if models have not learned invariant features or robust policies~\cite{xia2023generalization}. \rev{In our context here, we specifically refer to \emph{generalization} as the ability of a system to operate robustly across (i) different terrain types imposed by testing in new locations, (ii) different visual conditions caused by weather and lighting variations, and (iii) different embodiments when deployed on distinct robotic platforms. This framing aligns with the “Visual + Behavioral” axes of generalization described in~\cite{gao2025taxonomy}}.

The Earth Rover Challenge (ERC)~\cite{earthroverchallengeEarthRoverChallenge} serves as a rigorous testbed for evaluating precisely this kind of generalization. In this competition, identical sidewalk robots are deployed across \textbf{three continents and six countries}, \rev{spanning both urban and rural settings, and exposed to diverse real-world conditions, including variations in weather, and terrain (see Fig.~\ref{fig:erc_scenes}). Each robot is a low-cost, four-wheeled platform equipped only with noisy sensors, including a front and rear RGB camera, an IMU, wheel encoders, and a GPS unit with limited accuracy, while lacking any onboard computation. All perception and control must run remotely over a 4G link, where visual data arrives at only 3–5 Hz and inertial data below 10 Hz, introducing substantial delays and packet loss. Missions consist of reaching a sequence of GPS-defined checkpoints located hundreds of meters to several kilometers apart, with a difficulty tier that determines the score awarded. Partial credit is granted for completing a subset of checkpoints, and teams may use a limited number of brief tele-operated interventions, which reduce the mission score.

These stringent constraints, including low-cost sensing, slow and unreliable communication, and the absence of onboard processing, render traditional SLAM-based mapping and planning pipelines difficult to apply effectively. Instead, systems must robustly identify traversable areas and plan under uncertainty with minimal information, while tolerating severe distribution shift across environments. The inaugural 2024 edition underscored this difficulty: the top-performing autonomous system achieved only 36\% of the total possible score, far below the performance typically reported on standard academic benchmarks~\cite{earthroverchallengeEarthRoverChallenge,xiao2024erc}.}

To address these challenges, we propose GeNIE (Generalizable Navigation System for In-the-Wild Environments). An overview of GeNIE is shown in Fig. \ref{fig:system_overview}. At its core lies a robust traversability prediction model based on SAM2~\cite{kirillov2024sam2}, fine-tuned on a newly annotated dataset comprising 15,347 road scenes collected from diverse global environments. To efficiently produce training labels, we developed a semi-automatic annotation pipeline that combines SAM2-generated masks with lightweight human correction, significantly accelerating the annotation process while maintaining quality.

At inference time, the model outputs a pixel-wise traversability map, which is projected into a bird’s-eye-view cost map. From this map, random candidate paths are sampled and scored according to their traversability. However, naively selecting the path with the lowest cost often results in suboptimal behavior due to the presence of multiple, similarly scored paths in noisy or ambiguous regions. To mitigate this, we introduce a simple yet effective path fusion strategy that merges nearby paths meeting traversability constraints. The final navigation path is then selected from this fused set, improving both stability and robustness.

Our system was deployed in the ERC ICRA 2025 competition and achieved \textbf{79\% of the maximum score}, outperforming the second-best team by \textbf{17\%}.
In summary, our main contributions are as follows:
\begin{itemize}
    \item We present \textbf{GeNIE}, a generalizable navigation system that demonstrates state-of-the-art performance across deployments in six countries spanning three continents, under diverse terrain types, road styles, weather conditions, and lighting. GeNIE achieves the highest performance to date in these challenging real-world settings, setting a new benchmark for generalizable navigation.
    
    \item We introduce a geographically diverse traversability annotation dataset comprising 15,347 road images from 20 countries. To the best of our knowledge, this is the most geographically diverse dataset available for the traversability prediction task.
    
    \item We release a held-out benchmark for traversability prediction, consisting of data collected from 18 countries across 5 continents, featuring a wide variety of visual appearances and terrain styles. To the best of our knowledge, this is the first benchmark to evaluate traversability prediction under such diverse real-world scenarios.
    
    \item We have publicly released our codebase, pretrained model weights, annotated training dataset, and evaluation benchmark to support further research in real-world navigation.
\end{itemize}

\section{Related Work}

\subsection{Traversability Prediction}
Recent work has advanced traversability prediction using self-supervised and vision-based methods. WVN~\cite{frey23fast} and WayFASTER~\cite{gasparino24} learn terrain affordances directly from RGB images using self-supervision and short-horizon motion cues, enabling fast adaptation in unlabeled wild environments. Jeon et al.~\cite{jeon24} fuse geometry and appearance for vehicle-specific estimation, while Lambert et al.~\cite{lambert23} rely on proprioception-derived slip to label off-road terrain without human supervision.

Danesh et al.~\cite{danesh24} apply domain adaptation to transfer knowledge from simulation to real-world scenes. Probabilistic methods like~\cite{endo24} model uncertainty explicitly to improve safety on planetary rovers, and classical CNNs~\cite{dlee21} remain strong baselines in structured off-road terrains.

However, most existing methods do not generalize well to unfamiliar terrains or degraded visual conditions. In contrast, our method leverages the pre-trained vision foundation model SAM2~\cite{kirillov2024sam2} and a large, diverse dataset to achieve robust and generalizable traversability prediction, enabling deployment across drastically different settings without environment-specific retraining.

\subsection{Trajectory Planning with Path Fusion}

Sample-based planners like DWA~\cite{Fox1997DWA} implicitly perform path fusion by simulating velocity-parameterized rollouts and selecting those optimizing safety and goal progress. Elastic Band~\cite{Quinlan1993ElasticBand} and TEB~\cite{Rosmann2017TEB} refine a global path into smooth, tensioned corridors to mitigate local minima. Roll-out methods~\cite{Sten2025Rollout} simulate dynamics-aware trajectories in rough terrain, while BiC-MPPI~\cite{Jung2024BiCMPPI} clusters bidirectional rollouts to reduce switching and enhance goal convergence.

Trajectory-level clustering has also been used for motion prediction~\cite{Sung2012TrajCluster} and socially compliant navigation~\cite{Pruc2022FIPP}, demonstrating that grouping geometrically or behaviorally similar paths improves planning robustness. Silhouette-guided clustering~\cite{Rousseeuw1987Silhouette} is often used to tune the number of clusters, favoring precision in high-risk contexts.

Unlike these approaches, our method explicitly clusters candidate paths sampled from a BEV traversability map. We then fuse nearby paths and select the final trajectory based on heading alignment with the goal. This strategy mimics human navigation heuristics and enhances stability and success in long-horizon outdoor planning.

\section{Method}

\subsection{System Overview}
Given an input RGB image, our system first predicts pixel-wise traversability scores in the image space. The predicted traversable regions are then projected into a local 2D bird's-eye view (BEV) cost map. Multiple candidate paths are sampled from this BEV map. We then perform path fusion on the top-$K$ paths with the lowest traversability costs to generate more stable and consistent trajectories. Finally, the best fused path is selected based on its alignment with the goal location. An overview of the system pipeline is illustrated in Figure~\ref{fig:system_overview}.

\subsection{Traversability Estimation}
\label{sec:Traversability-estimation}
\begin{figure}[t]
    \centering
    \includegraphics[width=8.9cm]{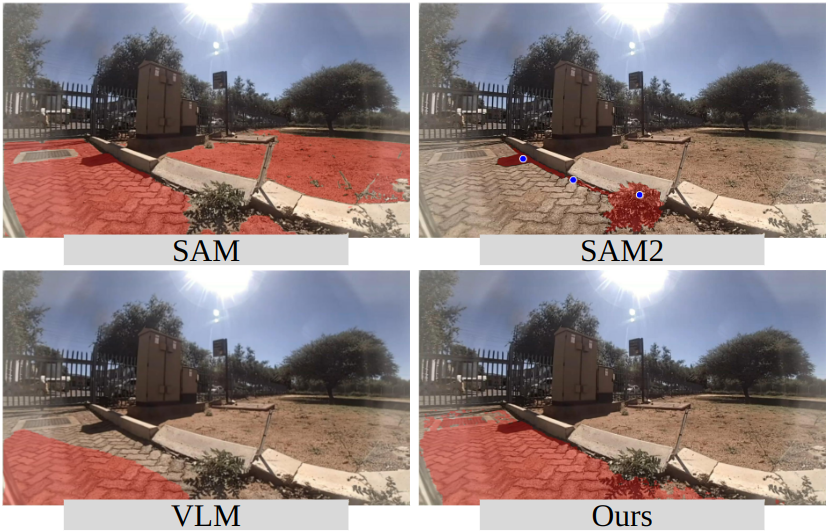}
    \caption{Comparison of traversability predictions (red regions) from SAM, SAM2, VLM (gemini-2.5-flash), and our proposed approach. For SAM, a text prompt describing navigable regions is used directly. Since SAM2 does not support text prompts, we first query the VLM (gemini-2.5-flash) to identify relevant points in the image, which are then used as prompts for SAM2.}
    \label{fig:Traversability_comparison}
    \vspace{-1em}
\end{figure}

To robustly estimate traversability across diverse environments, we build upon the pretrained SAM2 model \rev{(sam2.1 hiera tiny)}. However, naive application of SAM2 was inadequate for two main reasons. First, it tends to segment individual objects rather than the entire navigable surface, which often spans multiple visually similar regions. For example, in scenes where the flat ground is divided by markings or textures, SAM2 frequently outputs fragmented segments covering only parts of the traversable area. Second, SAM2 requires explicit point or bounding box prompts for inference and cannot directly generate masks based on the abstract concept of \emph{traversable area}. While models like SAM support text prompts, they perform poorly on traversability segmentation tasks (see Figure~\ref{fig:Traversability_comparison}).

Due to the lack of existing datasets for traversability segmentation across diverse terrains, we developed a semi-automatic annotation pipeline built on the FrodoBot dataset. Our data collection spans outdoor scenes from seven countries, covering a wide range of environments including grasslands, urban parks, deserts, and rural roads. To support scalable annotation, we used SAM2-generated region proposals as initial masks, which human annotators refined using a lightweight graphical interface. This process enabled the creation of 15,347 high-quality annotated frames with minimal manual effort.

A key feature of our dataset is that each traversable mask includes only regions that are directly navigable from the robot's current position, typically located near the bottom center of the image. This constraint ensures that the model learns to predict not just static free space, but the subset of the scene that is immediately reachable, thereby making the learned traversability more actionable for downstream planning (Figure~\ref{fig:Traversability_comparison})

To eliminate the reliance on manually provided prompts during inference, we removed the prompt encoder and instead directly optimized a learnable prompt token. Conceptually, this token encodes the notion of a ``traversable'' or ``navigable'' area. We refer to this modified version of SAM2 as \textbf{SAM-TP} (SAM for Traversability Prediction). All SAM-TP models are initialized with SAM2 pretrained weights~\cite{kirillov2024sam2} \rev{and trained on our annotated dataset using the Adam optimizer with a batch size of 20, a learning rate of $1\times10^{-6}$, and 10 epochs. Training was performed on a single A100 GPU and required less than 4 hours.}

\begin{itemize}
    \item \textbf{SAM-TP (D1)}: Fine-tunes only the last block of the two blocks in the mask decoder.
    \item \textbf{SAM-TP (D2)}: Fine-tunes all blocks of the mask decoder.
    \item \textbf{SAM-TP (E6)}: Fine-tunes all layers of the mask decoder and the last 6 of the 12 blocks in the image encoder.
    \item \textbf{SAM-TP (Ours)}: Fine-tunes all layers of both the mask decoder and the image encoder. 
\end{itemize}

To more thoroughly assess generalization capabilities, we collected an additional set of \rev{484} images across 17 countries, encompassing a wide range of terrain types, weather conditions, and lighting scenarios. All images in this benchmark were excluded from training and annotated using our semi-automatic pipeline. This dataset provides a  benchmark for evaluating model performance under realistic conditions. We organize the benchmark into three categories:

\begin{itemize}
    \item \textbf{Normal (N)}: Contains 167 scenes captured under typical conditions, including mild lighting and clearly defined paths. This set serves as the baseline for model performance.

    \item \textbf{Difficult (D)}: Comprises 160 scenes featuring challenging conditions commonly observed in the Earth Rover Challenge, such as uneven terrain, heavy rain, strong shadows, overexposure, occlusions, and visually fragmented surfaces caused by markings or textures. This set is designed to assess the model’s robustness to real-world variability.

    \item \textbf{Cross-embodiment (CE)}: Includes 157 scenes collected using a different robot platform (FrodoBot Mini), which differs significantly in height, aspect ratio, camera placement, and optical characteristics compared to the FrodoBot Zero used during training. This set evaluates the model’s ability to generalize across embodiments.
\end{itemize}

In addition to evaluating variants of the SAM-TP model, we also benchmarked several baselines using our proposed test set:

\begin{itemize}
    \item \textbf{WVN~\cite{frey23fast}}: The traversability prediction model introduced in~\cite{frey23fast}.
    
    \item \textbf{SAM~\cite{kirillov2023segment}}: The original SAM model prompted with a text description of the navigable region.
    
    \item \textbf{VLM gemini-2.5-flash~\cite{team2023gemini}}: A vision-language model, gemini-flash-2.5, which directly generates segmentation masks in response to textual queries.
    
    \item \textbf{SAM-TP (single scene)}: SAM-TP fine-tuned on a dataset collected from a single environment.
    
    \item \textbf{SAM-TP (20\%)}: SAM-TP fine-tuned using 20\% of the full training dataset.
    
    \item \textbf{SAM-TP (50\%)}: SAM-TP fine-tuned using 50\% of the full training dataset.
\end{itemize}

\begin{table}[t]
\centering
\begin{tabular}{
    >{\raggedright\arraybackslash}m{0.36\linewidth}
    >{\centering\arraybackslash}m{0.13\linewidth}
    >{\centering\arraybackslash}m{0.13\linewidth}
    >{\centering\arraybackslash}m{0.13\linewidth}
}
\toprule
\multirow{2}{*}{\textbf{Method}} & \multicolumn{3}{c}{IoU across three datasets $\uparrow$} \\
& \textbf{Normal} & \textbf{Difficult} & \textbf{CE} \\
\midrule
WVN~\cite{frey23fast} & 0.6157 & 0.4684 & 0.4551\\
SAM~\cite{kirillov2023segment} & 0.8759 & 0.7270 & 0.7610\\
VLM~\cite{team2023gemini} & 0.6528 & 0.5251 & 0.6206\\
\midrule
SAM-TP (single scene) & 0.8662 & 0.6867 & 0.8230\\
SAM-TP (20\% dataset) & 0.9120 & 0.7503 & 0.8889\\
SAM-TP (50\% dataset) & 0.9237 & 0.7758 & 0.9071\\
\midrule
SAM-TP (D1) & 0.8337 & 0.6572 & 0.8215\\
SAM-TP (D2) & 0.8480 & 0.6843 & 0.8375\\
SAM-TP (E6) & 0.9248 & 0.7974 & 0.9256\\
Ours & \textbf{0.9289} & \textbf{0.7989} & \textbf{0.9262}\\
\bottomrule
\end{tabular}
\caption{Zero-shot segmentation accuracy on our test dataset. The table reports the average Intersection over Union (IoU) across three evaluation conditions: normal, difficult, and cross-embodiment (CE). It compares baseline methods, our proposed approach trained with different dataset subsets, and various fine-tuning strategies. “SAM-TP (D1)” and “SAM-TP (D2)” denote partial fine-tuning of the mask decoder; “SAM-TP (E6)” includes additional encoder layers; and “Ours” fine-tunes the entire image encoder and mask decoder.}
\label{tab:benchmark_results}
\vspace{-1.5em}
\end{table}

The benchmark results are summarized in Table~\ref{tab:benchmark_results}. The \textbf{Difficult} test set presents significant challenges for all models, highlighting the complexity of real-world conditions. Among all methods, the proposed \textbf{SAM-TP} achieves the highest performance, demonstrating strong generalization across diverse environmental conditions and robotic embodiments.

\subsection{Bird's-Eye View Map Generation}
\label{sec:BEV}

Due to the lack of a depth camera on our robotic platform, we could not construct the bird's-eye view (BEV) map by directly projecting point clouds, as is commonly done. While several monocular depth estimation models exist~\cite{yang2024depth, bochkovskii2024depth}, we found that their predictions are insufficiently accurate for our application. Inaccurate depth estimates can result in critical failures such as collisions or the robot falling into a trench.

To overcome this, we adopt a simple yet effective approach that leverages the segmented traversable regions and a known camera height prior, which is a parameter typically available in robot systems. Instead of relying on the standard pinhole camera model, which fails under severe lens distortion, we use a generic camera model~\cite{schops2020having} to calibrate the camera. This calibration yields a directional ray vector $\hat{d}$ for each pixel in the image.

For any pixel classified as traversable, we assume the corresponding 3D point lies on the ground plane. Given the known camera height $h$, we solve for the scale $s$ in the equation: \rev{$s \hat{d}_y = -h$.}
This allows us to recover the 3D position of the pixel in the camera coordinate frame as $s \hat{d}$. These points are then projected into the BEV map.

\subsection{Path Planning and Fusion}
\label{sec:planning}

Given the BEV cost map, the robot must predict a safe and traversable path that ideally leads toward the goal. A common strategy is to use sample-based planning~\cite{frey23fast}, where candidate paths are sampled from the robot’s current position toward the goal—or toward intermediate goals if the final goal is far away. The optimal path is then selected by minimizing a cost function of the form:
\begin{equation}
C(l) = \sum_{i=1}^{n} f(p_i) + \beta \sum_{i=1}^{n} g(p_i)
\label{eq:cost}
\end{equation}
where $p_i$ is the $i$-th waypoint along the sampled path, $f(\cdot)$ is the cost derived from the traversability map, and $g(\cdot)$ is a goal-directed cost that encourages progress toward the target.

\begin{figure}[t]
    \centering
    \setlength{\tabcolsep}{4pt} 
    \renewcommand{\arraystretch}{0.5} 
    \begin{tabular}{cc}
        \includegraphics[width=0.45\linewidth]{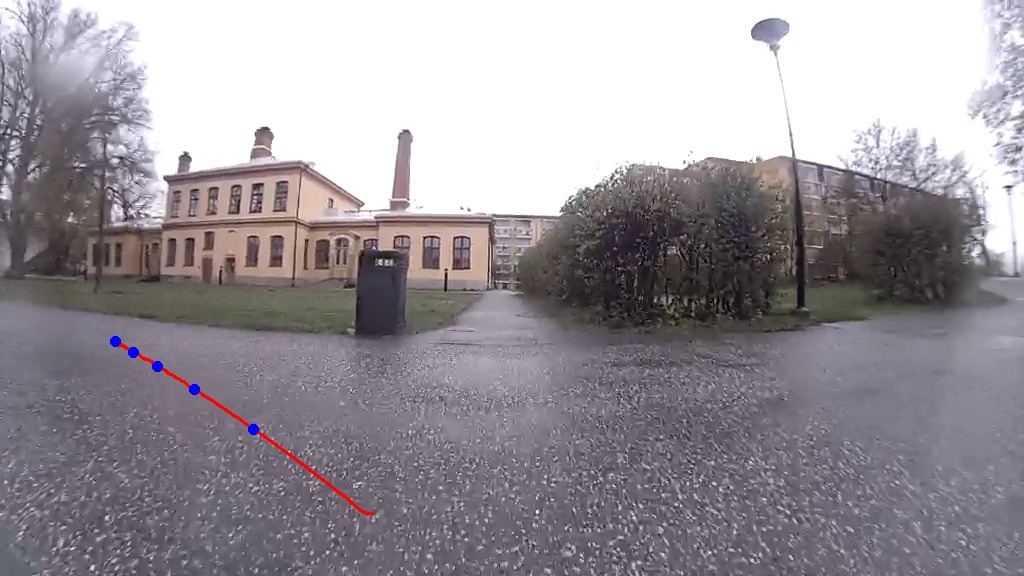} &
        \includegraphics[width=0.45\linewidth]{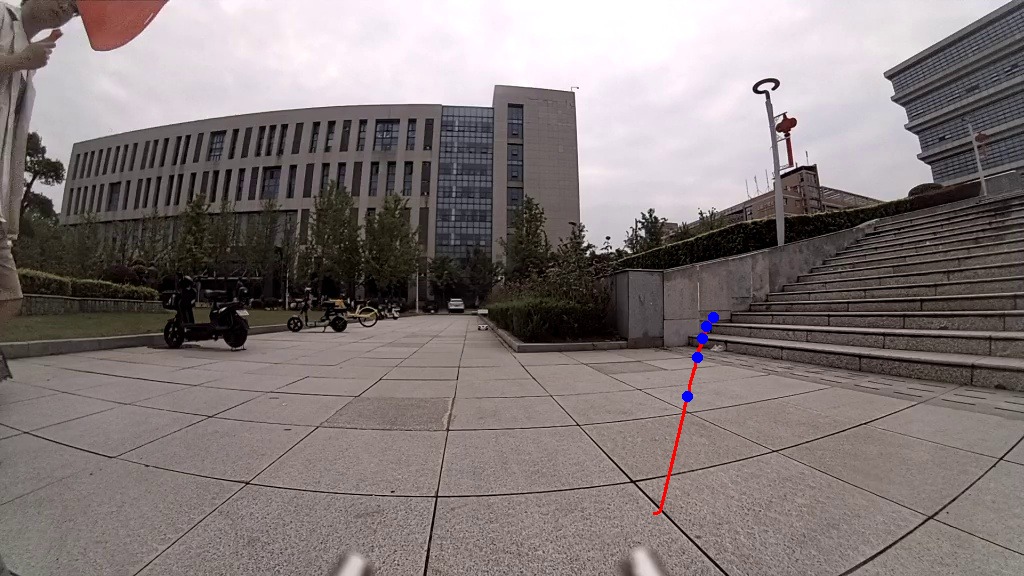} \\
        
        \includegraphics[width=0.45\linewidth]{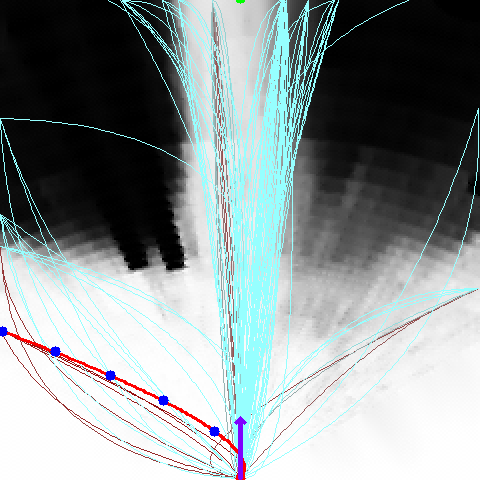} &
        \includegraphics[width=0.45\linewidth]{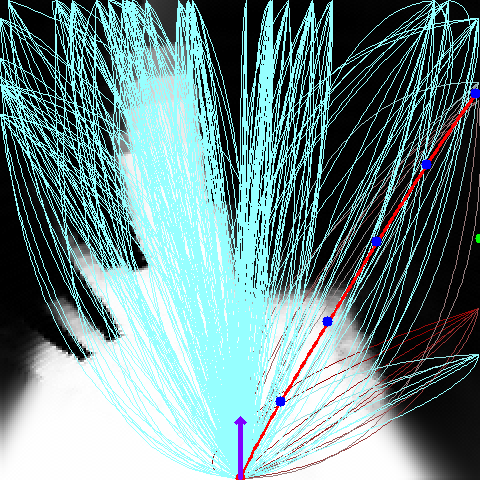} \\
        \small (a) No fusion ($\beta$ too small) & \small (b) No fusion ($\beta$ too large) \\[1ex]

        \includegraphics[width=0.45\linewidth]{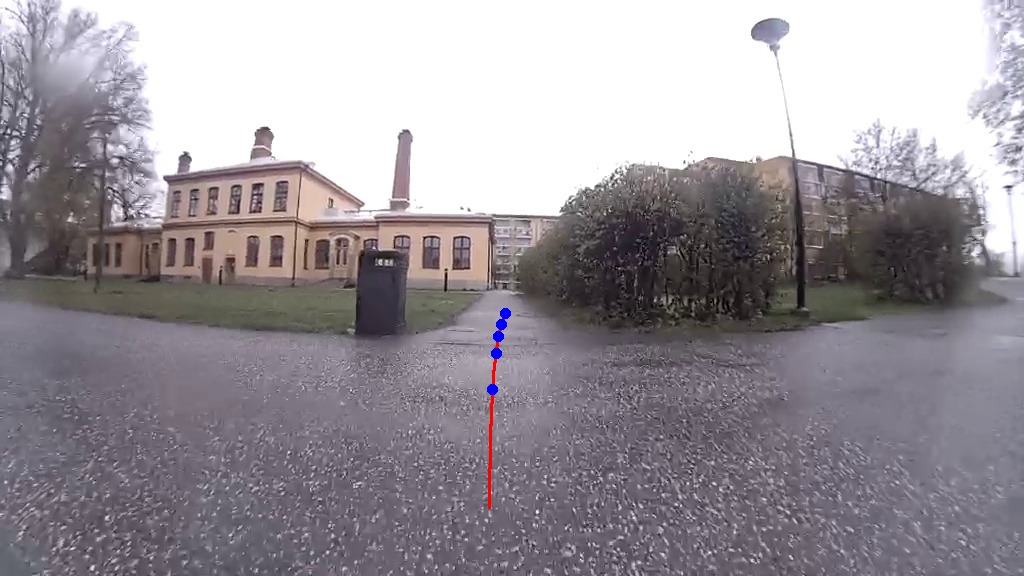} &
        \includegraphics[width=0.45\linewidth]{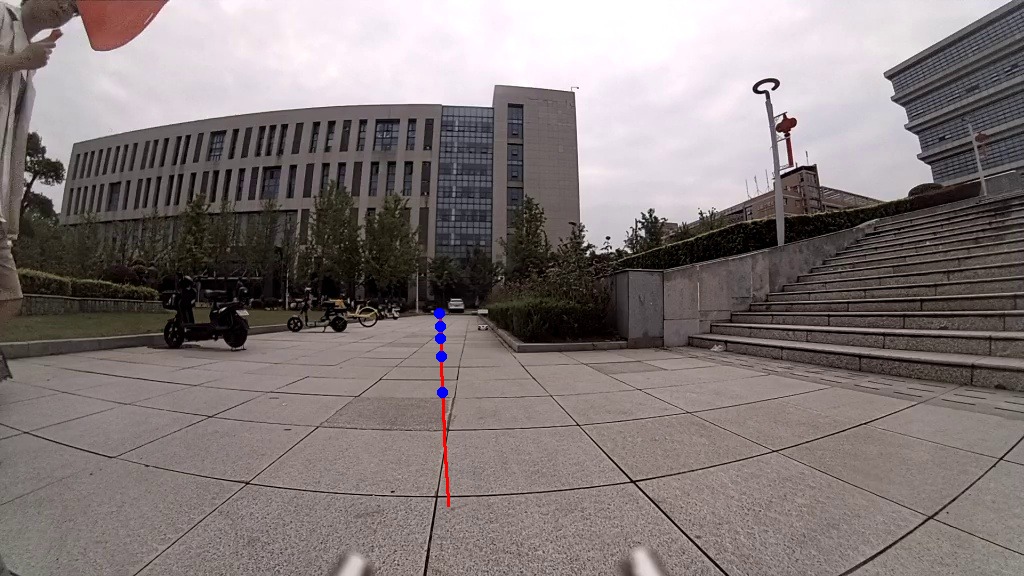} \\

        \includegraphics[width=0.45\linewidth]{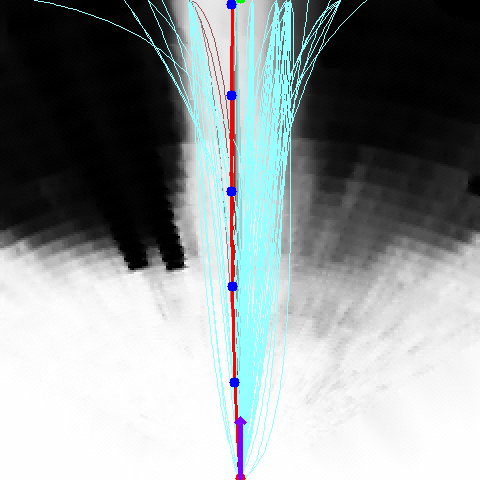} &
        \includegraphics[width=0.45\linewidth]{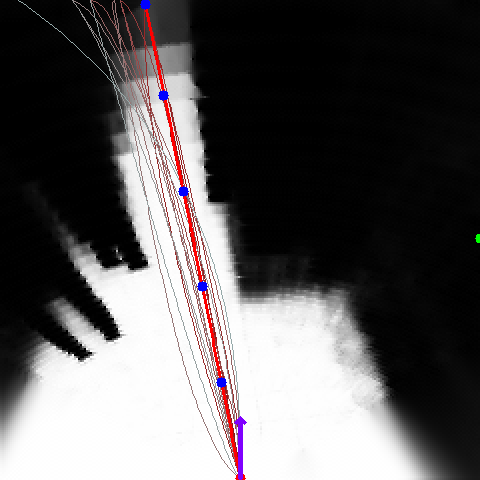} \\
        \small (c) With path fusion & \small (d) With path fusion \\
    \end{tabular}
\caption{Comparison of failure and success cases under different path planning strategies. The first and third rows show the robot’s current observation overlaid with the projected planned trajectory. The second and fourth rows display the BEV cost map with sampled paths; the selected path is shown in red, and the green dot at the edge of the map indicates the goal direction. Without path fusion, the planner tends to select paths with lower traversability when $\beta$ is small, or gets stuck in local minima when $\beta$ is large. In contrast, path fusion succeeds in both scenarios.}

    \label{fig:comparison_fusion_no_fusion}
    \vspace{-1em}
\end{figure}

However, this strategy often performs poorly when the goal lies hundreds of meters away. Two main issues arise:

\begin{enumerate}
    \item \textbf{Difficulty balancing traversability and goal costs}: When the goal-directed term $g(\cdot)$ dominates the cost function, the robot tends to move greedily toward the goal while ignoring environmental structure. This often results in myopic behavior and causes the robot to fall into local minima. Conversely, if the traversability cost $f(\cdot)$ dominates, the robot may ignore the goal and choose suboptimal paths that appear easier to traverse but lead in the wrong direction. In practice, there is no single value of $\beta$ that works robustly across all scenarios, and adaptively tuning $\beta$ at runtime is not feasible. Representative failure cases are shown in Figure~\ref{fig:comparison_fusion_no_fusion}.

    \item \textbf{Path switching due to cost equivalence}: In real-world deployments, robots typically employ a receding horizon strategy, continuously replanning as new observations become available. However, when many candidate paths have similar costs, the robot may frequently switch between them, leading to unstable behavior and slow overall progress.
\end{enumerate}

Tuning the weight $\beta$ between the goal-directed term and the traversability cost does not resolve these issues, as different situations require different trade-offs. Ideally, the robot should adaptively balance these objectives based on the current context. As shown in Figure~\ref{fig:pr_and_beta_accuracy}, the success rate of path selection under this naive strategy remains low across a wide range of $\beta$ values.

In contrast, humans rarely plan paths at the pixel level. Instead, we abstract navigable structure from visual inputs and reason at a higher level: along paths or tracks~\cite{Balaguer2016}. Whether following an urban road or an informal dirt path flanked by grass, we tend to follow the path until a junction appears, and only then do we make directional decisions. This behavior reflects a strong prior: if a path leads generally toward the goal, it is likely to be successful. This prior holds in many real-world outdoor settings.

To incorporate this prior, we propose a simple but effective planning method that operates at the path level. Given the BEV cost map, we first sample candidate paths from the robot’s current position to intermediate goals within its current field of view. These paths are parameterized as first- and second-order polynomials connecting the start and end points. We then filter these candidates using the BEV cost map and retain the top $N$ lowest-cost paths.

To detect the traversable paths, we apply \textbf{path fusion} to the top $N$ candidate paths. We first cluster these paths using adaptive $k$-means, where clustering is based on the Euclidean distance between corresponding waypoints across different paths. The number of clusters $k$ is determined by optimizing the silhouette loss, which balances intra-cluster compactness and inter-cluster separation.

The silhouette loss is defined as the negative of the average silhouette coefficient across all data points:

\begin{equation}
\mathcal{L}_{\text{silhouette}} = -\frac{1}{N} \sum_{i=1}^{N} \frac{b(i) - a(i)}{\max\{a(i), b(i)\}},
\label{eq:silhouette}
\end{equation}

where $a(i)$ is the average distance between sample $i$ and all other points in the same cluster, and $b(i)$ is the minimum average distance between sample $i$ and points in the nearest neighboring cluster. A higher silhouette coefficient indicates better clustering quality, and minimizing the negative value guides the model toward an optimal number of clusters.

Since minimizing silhouette loss tends to favor finer-grained clusters, we perform a post-processing step to merge any clusters whose centroids lie within a predefined distance threshold. The centroid of each merged group is used as the final representative path.

We define a \textit{false positive (FP) merge} as the erroneous merging of distinct paths that should remain separate, e.g., merging two paths split by an obstacle, potentially leading to unsafe planning. Conversely, a \textit{false negative (FN) merge} occurs when two similar paths are not merged despite being part of the same structure. While FPs can lead to critical failures (e.g., collisions), FNs simply preserve redundant safe paths. Hence, we prefer \textbf{precision over recall} in determining the merging threshold. A Precision–Recall (PR) curve illustrating this trade-off is shown in Figure~\ref{fig:pr_and_beta_accuracy} (a).

After path fusion, if only a single representative path remains, we select it as the final plan. If multiple paths remain, we choose the one most aligned with the goal direction.

\begin{table}[t]
\centering
\begin{tabular}{
    >{\raggedright\arraybackslash}m{0.32\linewidth}
    >{\centering\arraybackslash}m{0.16\linewidth}
    >{\centering\arraybackslash}m{0.16\linewidth}
    >{\centering\arraybackslash}m{0.16\linewidth}
}
\toprule
\textbf{Method} & \textbf{PIP (\%) $\uparrow$} & \textbf{PIR (\%) $\uparrow$}  & \textbf{PSA (\%) $\uparrow$} \\
\midrule
No Fusion ($\beta=0.1$)  & - & - & 27.3 \\
No Fusion ($\beta=1$)  & - & - & 42.9 \\
No Fusion ($\beta=10$)  & - & - & 36.4 \\
\midrule
k-means Only & 75.0 & \textbf{75.0} & 61.9 \\
Euclidean Selection  & \textbf{84.6} & 57.9 & 71.4 \\
Ours  & \textbf{84.6} & 57.9 & \textbf{85.7} \\
\bottomrule
\end{tabular}
\caption{Evaluation of path fusion and selection strategies with Path Identification Precision (PIP), Path Identification Recall (PIR), and Path Selection Accuracy (PSA).}
\label{tab:path_fusion_eval}
\vspace{-1em}
\end{table}

To evaluate the effectiveness of our path fusion and selection approach, we curated a test set consisting of road images with varying numbers of path branches, annotated by human raters. We use the following metrics in our empirical study:

\begin{figure}[t]
  \centering
  \setlength\tabcolsep{1em}     
  \begin{tabular}[t]{@{}c@{}c@{}}  
    
    \begin{tikzpicture}[baseline=(current bounding box.north)]
  \begin{axis}[
    width=0.48\columnwidth,
    height=0.45\columnwidth,
    xlabel={Recall},
    ylabel={Precision},
    xmin=0.25, xmax=1.0,
    ymin=0.25, ymax=1.0,
    grid=both,
    tick label style={font=\scriptsize},
    label style={font=\small},
    title={PR Curve},
    title style={font=\small},
  ]
    \addplot[thick,mark=*,mark size=1.2pt] coordinates {
      (0.32,   0.80)    
      (0.50,   0.9355)  
      (0.7377, 0.9375)  
      (0.9091, 0.9524)  
      (0.9697, 0.8889)  
      (0.9683, 0.8472)  
    };
    \node[anchor=west,font=\scriptsize] at (axis cs:0.32,   0.80)   {10};
    \node[anchor=west,font=\scriptsize] at (axis cs:0.50,   0.9355) {20};
    \node[anchor=west,font=\scriptsize] at (axis cs:0.7377, 0.9375) {30};
    \node[anchor=west,font=\scriptsize] at (axis cs:0.9091, 0.9524) {40};
    \node[anchor=west,font=\scriptsize] at (axis cs:0.9697, 0.8889) {50};
    \node[anchor=west,font=\scriptsize] at (axis cs:0.9683, 0.8472) {60};
  \end{axis}
\end{tikzpicture}

    &
    \begin{tikzpicture}[baseline=(current bounding box.north)]
      \begin{axis}[
        width=0.48\columnwidth,
        height=0.45\columnwidth,
        xlabel={$\beta$},
        ylabel={Accuracy},
        ymin=0.15, ymax=0.5,
        xtick={0,1,2,3,4,5,6},
        xticklabels={0.1,0.2,0.5,1.0,2.0,5.0,10.0},
        grid=both,
        tick label style={font=\scriptsize},
        label style={font=\small},
        title={Path Selection Accuracy},
        title style={font=\small},
      ]
        \addplot+[
          only marks,mark=*,mark size=2pt,
          error bars/.cd,y dir=both,y explicit,
          error bar style={line width=1pt},
          error mark options={rotate=90,mark size=4pt,line width=1pt},
        ] coordinates {
          (0,0.2727) +- (0,0.0297)
          (1,0.2273) +- (0,0.0372)
          (2,0.3182) +- (0,0.0321)
          (3,0.4091) +- (0,0.0295)
          (4,0.3636) +- (0,0.0241)
          (5,0.3636) +- (0,0.0341)
          (6,0.3636) +- (0,0.0247)
        };
      \end{axis}
    \end{tikzpicture}
    
    \\[\abovecaptionskip]
    \scriptsize (a) \rev{Precision–Recall}
    &
    \scriptsize (b) Accuracy vs.\,$\beta$
  \end{tabular}

  \caption{%
    (a) Precision–Recall curve for different merging  threshold for path identification
    (b) Path selection accuracy for No Fusion, as a function of~$\beta$. 
  }
  \label{fig:pr_and_beta_accuracy}
  \vspace{-1em}
\end{figure}

\begin{itemize}
    \item \textbf{Path Identification Precision (PIP):} The proportion of correctly merged path branches (true positives) out of all predicted path merges.
    
    \item \textbf{Path Identification Recall (PIR):} The proportion of correctly identified path branches (true positives) out of all ground truth two-branch cases.
    
    \item \textbf{Path Selection Accuracy (PSA):} The percentage of cases where the selected path matches the human-annotated correct branch.
\end{itemize}

We evaluate several variants of our pipeline:

\begin{itemize}
    \item \textbf{No Fusion (NF):} Directly selects the lowest-cost path from the sampled set using the cost function in Equation~\ref{eq:cost}.
    
    \item \textbf{k-Means Only (KM):} Clusters paths using k-means, without any post-processing or merging.
        
    \item \textbf{Euclidean Selection (ES):} Uses k-means + merge, then selects the path whose endpoint is closest (Euclidean distance) to the goal.
    
    \item \textbf{Angular Selection (Ours):} Uses k-means + merge, then selects the path with heading direction most aligned with the goal direction.
\end{itemize}

The results are summarized in Table~\ref{tab:path_fusion_eval}. The \textbf{no-fusion} strategy yields the lowest path selection accuracy, and tuning the weight $\beta$ does not improve performance, as shown in Figure~\ref{fig:pr_and_beta_accuracy}(b). While using k-means alone results in the highest path identification recall due to the adaptive k-means assigning more clusters, it suffers from low path identification precision and lower path selection accuracy compared to strategies that include a merging step.

Among the merging-based strategies, selecting the path whose heading direction is most aligned with the goal direction achieves the highest path selection accuracy. This result aligns with intuitive human path planning behavior, where a balance between feasibility and goal alignment is naturally considered. Some of the visualized path fusion results are shown in Figure~\ref{fig:comparison_fusion_no_fusion}

Putting everything together, our planning algorithm is summarized in Algorithm~\ref{alg:path_planning}. In our implementation, the random path set is generated only once at initialization, and only cost evaluations are updated during each planning cycle—making the approach computationally efficient.

\begin{algorithm}[t]
\caption{Path Planning with Path Fusion}
\label{alg:path_planning}
\KwIn{Random paths $\mathcal{T} = \{l_1, l_2, \ldots, l_M\}$, GPS goal $\mathbf{g}$, RGB image $I$, traversability model $\phi$}
\KwOut{Selected trajectory $l^*$}

$m = \text{ProjectToBEV}(\phi(I))$ \tcp*[l]{BEV cost map}

$\mathcal{T}_K \leftarrow \text{TopK}_{l_j \in \mathcal{T}} \left[ \sum_i m(p_i) \right]$ \tcp*[l]{Top-K paths}

$\mathcal{C} \leftarrow \text{AdaptiveKMeans}(\mathcal{T}_K)$ \tcp*[l]{Clustering}

$\mathcal{C}_\text{merged} \leftarrow \text{MergeCloseCentroids}(\mathcal{C})$ 

\If{$|\mathcal{C}_\text{merged}| = 1$}{
    $l^* \leftarrow \text{Centroid}(\mathcal{C}_\text{merged})$ 
}
\Else{
    $l^* \leftarrow \arg\min_{l \in \mathcal{C}_\text{merged}} \angle(l, \mathbf{g})$ 
}
\Return $l^*$

\end{algorithm}

\subsection{Closed-loop Control}

\rev{The control module implements a simple waypoint-tracking policy that prioritizes robustness over efficiency. Our system operates in a receding-horizon loop: at each timestep, the latest observation is used to replan a path via Algorithm~\ref{alg:path_planning}, and the robot then executes the trajectory toward the first waypoint before repeating the process with updated input. A collision-avoidance mechanism runs continuously in parallel, monitoring the predicted traversability map from each incoming frame to halt motion if a nearby hazard is detected. Concretely, after each planning step the robot aligns its heading with the next waypoint, and then advances by 1 m, unless the collision-avoidance module signals low traversability in the near-front region. Since our SAM-TP model runs at 10 Hz on an RTX 3090 GPU while frames arrive at only 3–5 Hz, collision checks can be executed in real time without adding latency. The planning module itself runs at approximately 5 Hz. Although this synchronous replan–align–move cycle is less efficient than continuous controllers such as DWA, it was deliberately chosen for robustness under the high-latency conditions of ERC. }

\rev{In some cases, the planner may guide the robot into dead ends such as dense grass, enclosed areas, or regions near obstacles, where it fails to find a valid path. To handle these situations, we introduce a vision-language model (VLM)-assisted recovery module. Given a short history of recent observations, the VLM is prompted with a natural language query on how to return to the previously navigable route. This allows the robot to recover from local failures using high-level semantic reasoning. Please see the online supplementary material for further details.}


\section{ERC Competition Results}

The GeNIE system was deployed in the 2025 Earth Rover Challenge (ERC) across Africa, South America, and North America. In each mission, the robot, equipped with front and rear RGB cameras, IMU, wheel encoders, and GPS, was tasked with reaching a sequence of GPS-defined checkpoints over distances of several hundred meters to a few kilometers. Scoring was based on mission difficulty and the number of checkpoints completed, with penalties applied for human interventions, and a score of zero if the robot flipped over. The overall competition setup and scoring rules are described in Section~\ref{sec:introduction} and in~\cite{earthroverchallengeEarthRoverChallenge}.

We report the following evaluation metrics:
\begin{itemize}
    \item \textbf{Total Score}: The cumulative score earned by the team, calculated according to the scoring rules described above.
    
    \item \textbf{Normalized Score}: The total score divided by the maximum possible score (30), providing a percentage-based measure of performance.
    
    \item \textbf{Interventions}: The total number of human interventions used across all missions in which the team score was non-zero\footnote {\rev{This differs from the official count which is total number of interventions across all missions, regardless of score. This better reflects the model's ability to navigate since zero scores indicate a non-recoverable failure (e.g., the robot flipping over).}}
\end{itemize}

Our system achieved \textbf{79\%} of the maximum possible score, outperforming the second-place team by 17\%. Full results are provided in Table~\ref{tab:challenge_leaderboard}. Notably, our system required zero human interventions, demonstrating its robustness and adaptability across challenging and varied environments.

\begin{table}[t]
\centering
\begin{tabular}{lccc}
\toprule
\textbf{Team} & \textbf{Total Score $\uparrow$} & \textbf{Norm. Score $\uparrow$} & \textbf{Interventions $\downarrow$} \\
\midrule
Human Player & 30.00 & 1.00 & -  \\
\midrule
GeNIE (ours) & \textbf{23.78} & \textbf{0.79} & \textbf{0}  \\
Team A & 18.74 & 0.62 & 1  \\
Team B & 8.07 & 0.27 & 11  \\
Team C & 7.39 & 0.25 & 2  \\
Team D & 5.89 & 0.20 & 6  \\
\bottomrule
\end{tabular}
\caption{Leaderboard summary of the top 5 teams from the challenge. The normalized score is computed relative to the maximum possible score of 30.}
\label{tab:challenge_leaderboard}
\vspace{-1em}
\end{table}

\section{Conclusion and Limitations}

In this paper, we present \textbf{GeNIE}, a navigation system capable of generalizing across a wide range of real-world environments. The key to this generalization lies in two components: (1) a robust traversability prediction model, and (2) an effective path planning strategy based on path fusion. Our system demonstrated strong performance in the Earth Rover Challenge (ERC), achieving first place across deployments in six countries.

GeNIE is not without limitations. \rev{The system failed to complete three missions during ERC: one mission resulted in a zero score due to a local minimum problem, and two missions yielded only partial scores due to inefficient control under high latency.

The local minimum problem arises because the system relies on single-frame traversability predictions without maintaining a local map, which limits its ability to reason over longer horizons. In Mission 9, for example, the robot refused to enter a passage with low predicted traversability, not realizing it was the only viable route to the target. Such cases suggest that effective traversability prediction may require persistent memory and long-term reasoning, enabling the robot to integrate past observations and replan accordingly.

A second limitation is the conservative control strategy adopted to ensure robustness under high-latency conditions. The robot executes commands in a sequential “observe–plan–execute–wait” cycle rather than continuously replanning. While this avoids overshooting behaviors, it reduces efficiency and caused timeouts in two missions. Incorporating more latency-aware or hybrid control strategies could improve efficiency without compromising robustness.

Finally, our BEV projection assumes a locally flat ground plane, which is generally valid within the 8 m planning horizon but may introduce errors on steep slopes. Although this did not cause failures in ERC, integrating depth sensing or monocular depth estimation could further improve reliability. Another limitation is that the current representation uses a single traversability token, which may be insufficient for robots with fundamentally different embodiments (e.g., legged versus wheeled).

Addressing these limitations by incorporating spatial memory for long-horizon reasoning, improving control strategies under latency, and relaxing the flat-ground assumption represents an important direction for future work.}

\section*{Acknowledgements}

This research / project is supported by A*STAR under its National Robotics Programme (NRP) (Award M23NBK0053).

\balance
\bibliographystyle{IEEEtran}  

\bibliography{references}

\end{document}